\documentclass[10pt,twocolumn,letterpaper]{article}

 \usepackage{cvpr}              

\usepackage[dvipsnames]{xcolor}

\definecolor{cvprblue}{rgb}{0.21,0.49,0.74}
\usepackage{tikz}
\newcommand*\circled[1]{\tikz[baseline=(char.base)]{
            \node[shape=circle,draw,inner sep=1pt] (char) {#1};}}
\usepackage[pagebackref,breaklinks,colorlinks,citecolor=cvprblue]{hyperref}

\def\paperID{*****} 
\def\confName{CVPR}
\def\confYear{2024}

\title{CARLOR @ Ego4D Step Grounding Challenge: Bayesian temporal-order priors for test time refinement}

\author{Carlos Plou*\\
University of Zaragoza\\
{\tt\small c.plou@unizar.es}
\and
Lorenzo Mur-Labadia*\\
University of Zaragoza\\
{\tt\small lmur@unizar.es}
\and
Ruben Martinez-Cantin\\
University of Zaragoza\\
{\tt\small rmcantin@unizar.es}
\and
Ana C.Murillo\\
University of Zaragoza\\
{\tt\small acm@unizar.es} 
\thanks{This work was supported by two DGA scholarships, project DGA T45\_23R, and MCIN/AEI/ERDF/NextGenerationEU/PRTR projects PID2021-125514NB-I00, PID2021-125209OB-I00 and TED2021-129410B-I00.}
}

\begin{document}
\maketitle
\begin{abstract}

The goal of the Step Grounding task is to locate temporal boundaries of activities based on natural language descriptions. This technical report introduces a Bayesian-VSLNet to address the challenge of identifying such temporal segments in lengthy, untrimmed egocentric videos. Our model significantly improves upon traditional models by incorporating a novel Bayesian temporal-order prior during inference, enhancing the accuracy of moment predictions. This prior adjusts for cyclic and repetitive actions within videos. Our evaluations demonstrate superior performance over existing methods, achieving state-of-the-art results on the Ego4D Goal-Step dataset with a 35.18 Recall Top-1 at 0.3 IoU and 20.48 Recall Top-1 at 0.5 IoU on the test set.

\end{abstract}    
\section{Introduction}
\label{sec:intro}

The step grounding (SG) task, introduced on the Ego4D Goal-Step extension \cite{song2024ego4d}, aims to localize the temporal boundaries (\texttt{start time}, \texttt{end time}) of an activity described in a free natural language expression within long, untrimmed egocentric videos as Figure \ref{fig:teaser} shows. For instance, it can enable students to query specific segments of a demonstration video or assist visually impaired individuals through wearable devices that provide real-time audio descriptions of their surroundings. 

\begin{figure}[!t]
    \centering
    \label{fig:teaser}
    \includegraphics[width=\columnwidth]{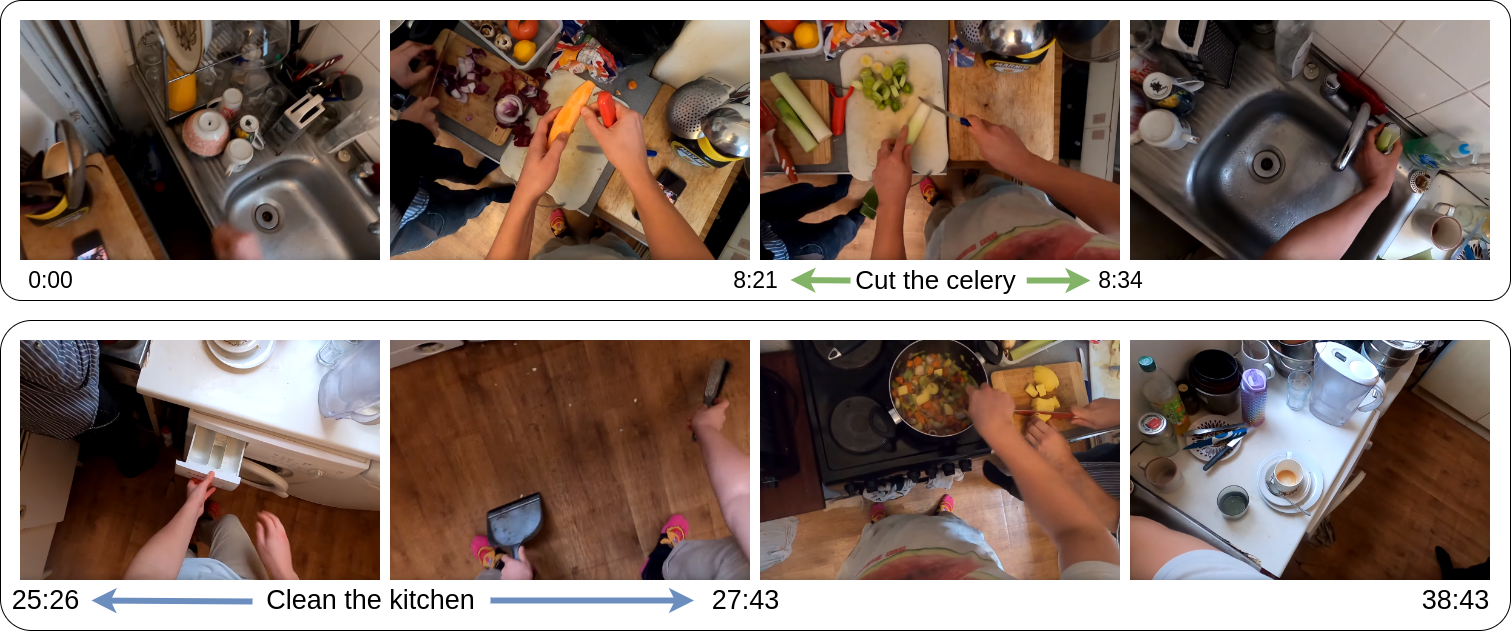}
    \caption{\textbf{Step grounding task:} given a long untrimmed video of a high-level procedural activity \textit{Make a soup}, the goal of the step grounding task is to localize the segment in the video that represents the  free-form natural language description of the step.}
    \label{fig:teaser}
\end{figure}

In recent years, the arrival of extensive narrated egocentric video datasets, such as Ego4D \cite{grauman2022ego4d}, introduced the Natural Language Queries (NLQ) challenge for episodic memory, whose goal is to identify video segments that answer specific queries such as \textit{"Where did I leave my keys?"}. This motivated several  works that attempt to localize text queries within videos.
GroundNLQ \cite{hou2023groundnlq} significantly advances this field with its two-stage pre-training pipeline designed for robust feature extraction, complemented by a multi-scale text-video transformer encoder that predicts segments at various levels. 
Simultaneously, CONE \cite{hou2022efficient} employs a sliding window technique to pre-filter candidate windows first and then refines 
the results with fine-grained intra-window proposals through Moment-DETR \cite{lei2021detecting}. Meanwhile, EgoEnv \cite{nagarajan2024egoenv} introduces an innovative approach by grounding videos in their 3D physical environments, rather than relying on naive temporal feature aggregation. 
Narrations-as-Queries \cite{ramakrishnan2023naq} transforms standard video-text narrations into training data for video query localization models, substantially boosting the performance across top models. Lastly, InternVideo \cite{wang2022internvideo} is a foundational model for video analysis
that achieved the state-of-the-art in various video-related challenging tasks.

\begin{figure*}[!t]
    \centering
\includegraphics[width=\linewidth]{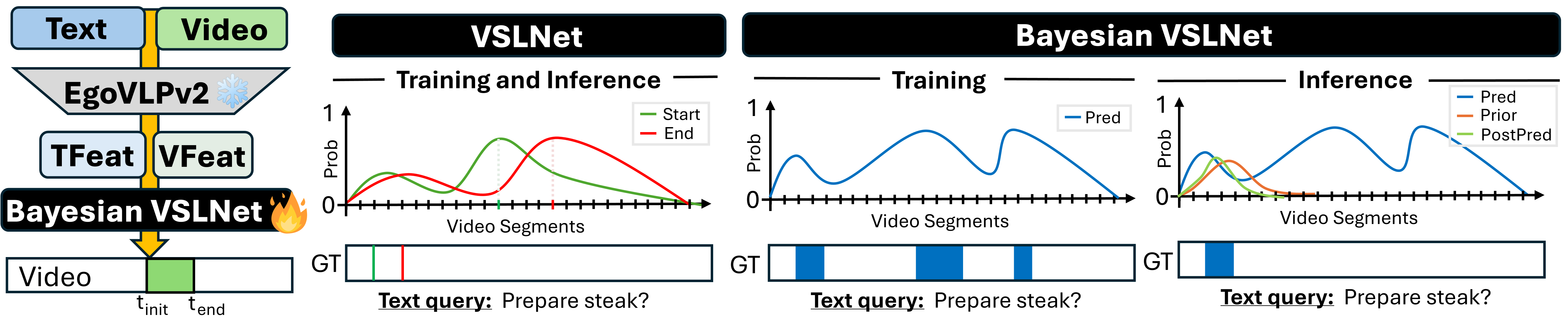}
    \caption{\textbf{Bayesian VSLNet.} We introduce two novel components: a novel head predicts the probability of the text query in each video segment and a Bayesian temporar-order prior refines the predictions during the inference stage.}
    \label{fig:model}
\end{figure*}

Despite its similarity to the NLQ challenge, the step grounding task incorporates additional difficulties: cyclic actions and extreme long videos.

First, the procedural content of the Ego4D-SG videos includes repeated and cyclic interactions which complicates the NLQ-training as it is currently formulated.
For example, \textit{"Rolling out the dough"} may occur repeatedly in a bakery shop scenario, spanning across the entire video. Unlike NLQ methods that rely on clear start and end labels, the repetitive nature of such actions introduces ambiguities that disrupt the training process when using the annotations provided. For the same text query, we are indicating that the action occurs in different moments, thus penalizing positive predictions due to a labelling problem.

Second, the increased length of the Ego4D-SG videos ($\mu=$1,516 seconds, with some videos extending over a few hours) compared to Ego4D-NLQ videos ($\mu=$495 seconds) exacerbates the ``needle in the haystack'' problem. The long duration of the videos dilutes the fine-grained correspondence (objects and scene distribution) between each video frame and the textual query due to interference from irrelevant frames, leading to a loss of contextual details. Approaches such as sliding-window techniques maintain temporal resolution but generate a high number of potential moment candidates, and contain insufficient global contextual information. The other alternative, sparse-sampling techniques, lose fine-grained information by averaging the frames.

\section{Methodology}


Our approach 
attempts to tackle the two challenges mentioned above, as  Figure \ref{fig:model} illustrates: cyclic actions and significantly longer duration of the videos.
First, we introduce a novel head that predicts for each video segment the probability of being linked with a given text query, instead of the start and end time of the action. This mitigates the disturbed effect of the provided cyclic and repeated actions during training. Next, we introduce a test time refinement strategy that exploits Bayes rule by adding a temporal-order prior to the predictions. This enables us to accurately predict every moment when the step appears in the video and focus the prediction according to the step order in the sequence.



\label{sec:methods}


\subsection{Text and Video Representations}
  Let us define $\mathcal{V}=\{\mathcal{V}_{i}\}_{i=1}^{N}$ as the set of videos. Each video $\mathcal{V}_{i}$ is composed by a set of ${N_{i}}$ frames (30 fps).
  Besides, each video $\mathcal{V}_{i}$ is linked with a set of text queries $\mathcal{T}_{i}=\{t_{i}^{j}\}_{j=1}^{m_{i}}$. 
  Thus, we first encode both videos $\mathcal{V}_{i}$ and text queries $\mathcal{T}_{i}$ in some representative features. 
  Video features $\mathcal{F}_{i}^{v}$ are extracted with a stride of $s=16$ frames -i.e., 1.875 features per second- and a sliding window of $32$ frames. Thus, for each video,  we get $n_{i} = \lfloor \frac{N_{i}}{s} \rfloor $ features $f^{v}_{ij}$.
  We extract the text query representation $\mathcal{F}_{i}^{t}$ with a pretrained LLM  model, obtaining a feature vector $f^{t}_{ij}$ per each step description $t_{i}^{j}$.
  
  

\subsection{VSLNet}
The original version of VSLNet (Video Span Localizing Network) \cite{zhang2020span} adopts an sparse sampling technique for compressing the video features $\mathcal{F}_{i}^{v}$ in $S$ segments. 
Let us define the final size of features $\mathcal{F}^{v}_{i}$ as $S_i=\min(n_{i}, S)$. In this context, each text query $t_{i}^{j}$ has two ground truth one-hot vectors $\textbf{s}_{ij}=\{s_{ij}^{k}\}_{k=1}^{S_i}, \textbf{e}_{ij}=\{e_{ij}^{k} \}_{k=1}^{S_i}$  where $k^{s}_{ij} = \arg \max_{k} s_{ij}^{k}$ and $k^{e}_{ij} = \arg \max_{k} e_{ij}^{k}$ are the indexes of the starting and the ending segment, respectively.

Afterwards, we project both visual $\mathcal{F}_{i}^{v}$ and text $\mathcal{F}_{i}^{t}$ features into the same latent space of dimension $d$ by way of two linear layers.
Then, a shared feature encoder formed by four convolutional layers followed by multi-head attention extracts contextual awareness and refines the representation.
Then, a context-query attention extracts cross-modal interaction features between the refined visual and text representations. The interaction is obtained by computing the context-to-query and query-to-context similarity matrices, indicating which are the most correlated parts, and aggregating the information between both modalities. This obtains a feature per segment with the content of the video refined with its significance in the query.

Finally, two LSTM layers aggregate these features to predict for each pair $V_{i}-t_{i}^{j}$ a probability score of starting and ending $(\mathbf{\hat{s}}_{ij},\mathbf{\hat{e}}_{ij})$ the given text query in each of the $S_i$ segments.
In this way, the model is trained under the assumption that each text query is repeated only once. This property prevents us from accurately modelling the scenario in which a specific text query is repeated along the video.


\subsection{Bayesian VSLNet}
 \paragraph{Training:} We group the identical text queries of a video and transform the ground truth into an ``event'' vector $\mathbf{p}_{ij}=\{p_{ij}^{k}\}_{k=1}^{S_i} \in \{0, 1\}^{S_i}$ that contains all the times that a given text query $t_{i}^{j}$ occurs in the video $\mathcal{V}_{i}$. In mathematical terms, 
 $$
 p_{ij}^{k} = \begin{cases}
    1, \text{if} \hspace{1mm} \exists  t_{i}^{q} \in \mathcal{T}_{i} \hspace{1mm} \text{s.t.}
    \begin{array}{l}
    \circled{1} \hspace{1mm} t_{i}^{q}=t_{i}^{j} \\
    \circled{2} \hspace{1mm} k^{s}_{iq} \leq k \leq  k^{e}_{iq}
    \end{array} \\
    0, \text{otherwise}
 \end{cases}
 $$
We substitute the VLSNet head by a unique LSTM layer followed by a sigmoid function that predicts $\mathbf{\hat{p}}_{ij} \in \left[0, 1\right]^{S_i}$ from the refined features. We train the model with a Binary Cross Entropy (BCE) loss.

\begin{figure}
     \centering
     \includegraphics[width=\columnwidth]{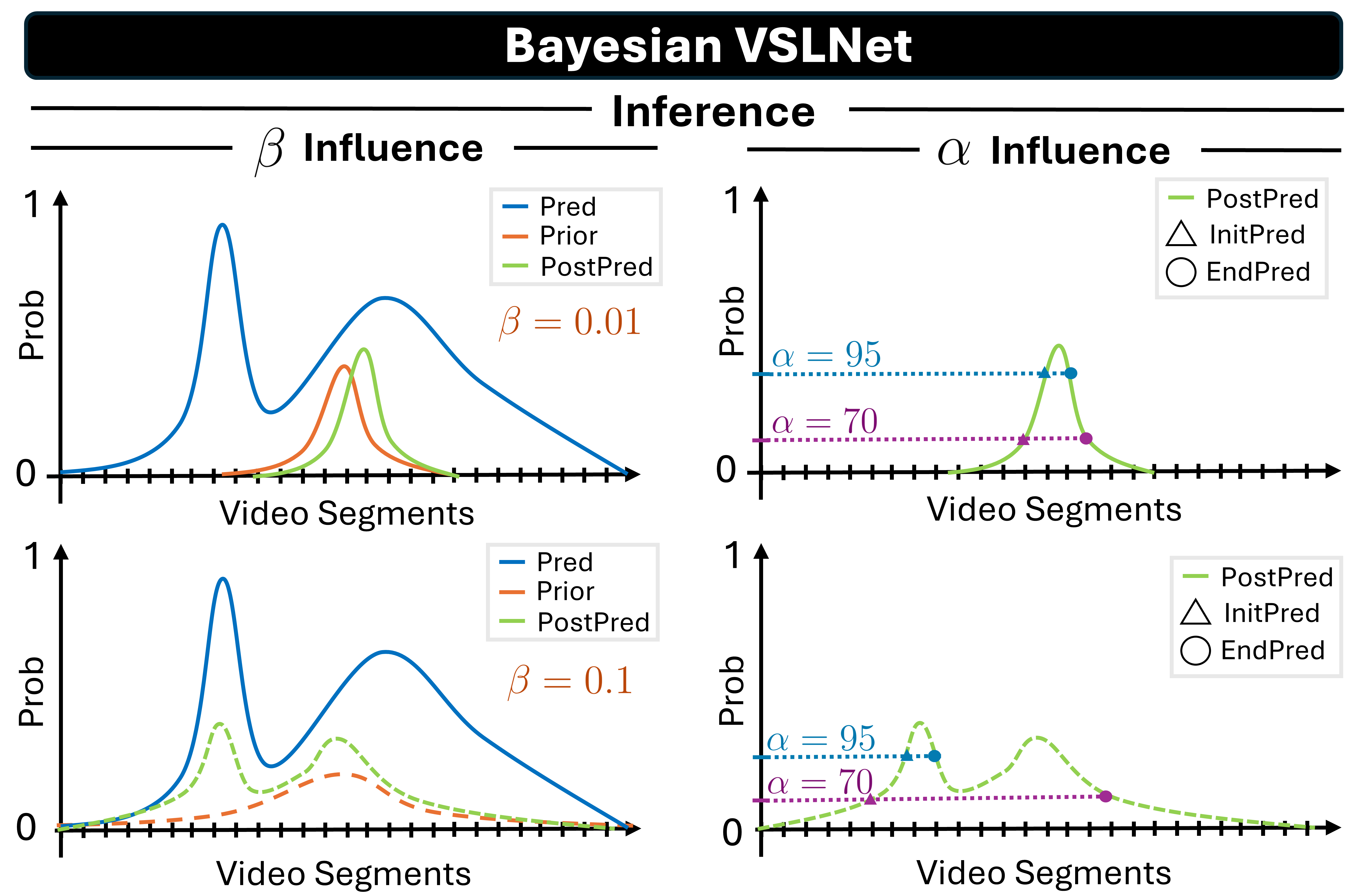}
     \caption{\textbf{Influence of the $\alpha$ and $\beta$ parameters at the inference stage.} $\alpha$ sets the threshold ($\alpha$-percentile of the posterior probability value $p^k_{ij}$) that controls the length of the segment. $\beta$ determines the variance of the prior that influences the posterior.}
     \label{fig:alphabeta}
 \end{figure}

 \paragraph{Inference:} We predict the starting and ending segment $\hat{k}^{s}_{ij}, \hat{k}^{e}_{ij}$ from the event probability distribution
$\mathbf{\hat{p}}_{ij}$. To achieve this goal, we search the most likely segment, that is, $k^{*}_{ij}= \arg \max_{k} \hat{p}_{ij}^{k}$ and, from it, we extend the segment forward $(k^{*}_{ij} \rightarrow k)$ and backward $(k \leftarrow k^{*}_{ij})$ until $\hat{p}^{k}_{ij}$ is under the $\alpha$-percentile $\mathbf{\hat{p}}^\alpha_{ij}$, where $\alpha$ controls the segment amplitude, as Figure \ref{fig:alphabeta} shows. In other terms,
 \begin{eqnarray}
    \hat{k}^{s}_{ij}: k \leq k^{*}_{ij}  \hspace{1mm}\text{s.t.}  & \hat{p}_{ij}^{k-1} < \hat{p}^\alpha_{ij}   \leq \hat{p}_{ij}^{k}, \\
    \hat{k}^{e}_{ij}:  k \geq k^{*}_{ij} \hspace{1mm} \text{s.t.}  & \hat{p}_{ij}^{k} \geq \hat{p}^\alpha_{ij} > \hat{p}_{ij}^{k+1}.
 \end{eqnarray}

 However, let us assume that a given text description is repeated $m$ times in a video. Each text query of them will have identical predictions $\mathbf{\hat{p}}_{ij}$, resulting in a unique segment prediction $(\hat{k}^{s}_{ij}, \hat{k}^{e}_{ij})$, loosing the information that this text query has occurred $m$ times at different moments. To mitigate this drawback, we exploit Bayes rule, adding a temporal-order prior $\mathbf{q}_{ij}=\{ q_{ij}^{k} \}_{k=1}^{S_i}$ to our predictions $\mathbf{\hat{p}}_{ij}$. We define this prior as the next Gaussian distribution, 
 \begin{equation}
   q_{ij}^{k} := \mathcal{N}\left(k; \frac{j \cdot S_i}{m_{i}}, S_i \cdot \beta \right),
 \end{equation}
 where $\beta$ controls the prior shape distribution and therefore its influence on the final posterior, as Figure \ref{fig:alphabeta} shows. Hence, we redefine our final predictions as
 \begin{equation}
   \hat{p}_{ij}^{k} :=  \frac{\hat{p}_{ij}^{k} \cdot q_{ij}^{k}}{\max(\mathbf{q}_{ij})}, 
 \end{equation}

\section{Results}

\begin{table}
    \small
    \centering
    \begin{tabular}{ccccccc}
    \toprule
         \multicolumn{2}{c}{Video Features} & \multicolumn{2}{c}{Text Features} & Conf. & \multicolumn{2}{c}{Val-IoU} \\
         \cline{1-2} \cline{3-4} \cline{5-5} \cline{6-7}  
         Omn. & VLPv2 & Bert & VLPv2 & S & 0.3 & 0.5 \\
         \toprule
         \checkmark & - & \checkmark & - & 128 & 11.77 & 7.77 \\
         - & \checkmark & - & \checkmark & 128 & 13.13 & 8.75 \\
         \checkmark & \checkmark & - & \checkmark & 512 & \textbf{16.26} & \textbf{11.81} \\
         \checkmark & \checkmark & - & \checkmark & 1024 & 15.28 & 11.18 \\
    \bottomrule
    \end{tabular}
    \caption{Results for different video and text features. We measure \texttt{r@1IoU0.3} and  \texttt{r@1IoU0.5} in the validation set.}
    \label{tab:features}
\end{table}

\begin{table}[]
\begin{tabular}{ccccc}
\hline
Model          & \multicolumn{2}{c}{Val-IoU} & \multicolumn{2}{c}{Test-IoU}    \\ \hline
Name           & 0.3          & 0.5          & 0.3            & 0.5            \\ \hline
VSLNet         & 16.26        & 11.81        & -              & -              \\
BayesianVSLNet & 18.15        & 8.97         & \textbf{35.18} & \textbf{20.48} \\ \hline
\end{tabular}
\caption{Results for the different modifications and configurations tested in our Bayesian VSLNet. We measure \texttt{r@1IoU0.3} and  \texttt{r@1IoU0.5} in the validation and test set.}
\end{table}

By default, VSLNet leverages Omnivore-L to obtain video features $\mathcal{F}_{i}^{v}$ and pretrained BERT \cite{devlin2018bert} model to get text query features $\mathcal{F}_{i}^{t}$. Following \cite{hou2022efficient}, we concatenate two feature extractors (Omnivore-L and EgoVLPv2) to obtain a richer video representation. We adopt the dual-encoder configuration of EgoVLPv2, and we extract the class token before the video projection. 

As it is observed in Table~\ref{tab:features}, VSLNet benefits from this alignment of visual features. From here on, we combined both visual features in all our experiments. We also increase the number of sampled segments to avoid the information loss due to the diffusion of individual video features in the segment. As the results show, the best configuration with 512 segments reports 16.26 Recall Top-1 at 0.3 IoU and 11.81 Recall Top-1 at 0.5 IoU.
Then, we adopt that configuration in our Bayesian VSLNet, with $\beta=0.1$ and $\alpha=85$ as hyper-parameters. The model achieves a 35.18 Recall Top-1 at 0.3 IoU and 20.48 Recall Top-1 at 0.5 IoU on the test set.

\begin{figure*}[t]
 \includegraphics[width=\textwidth]{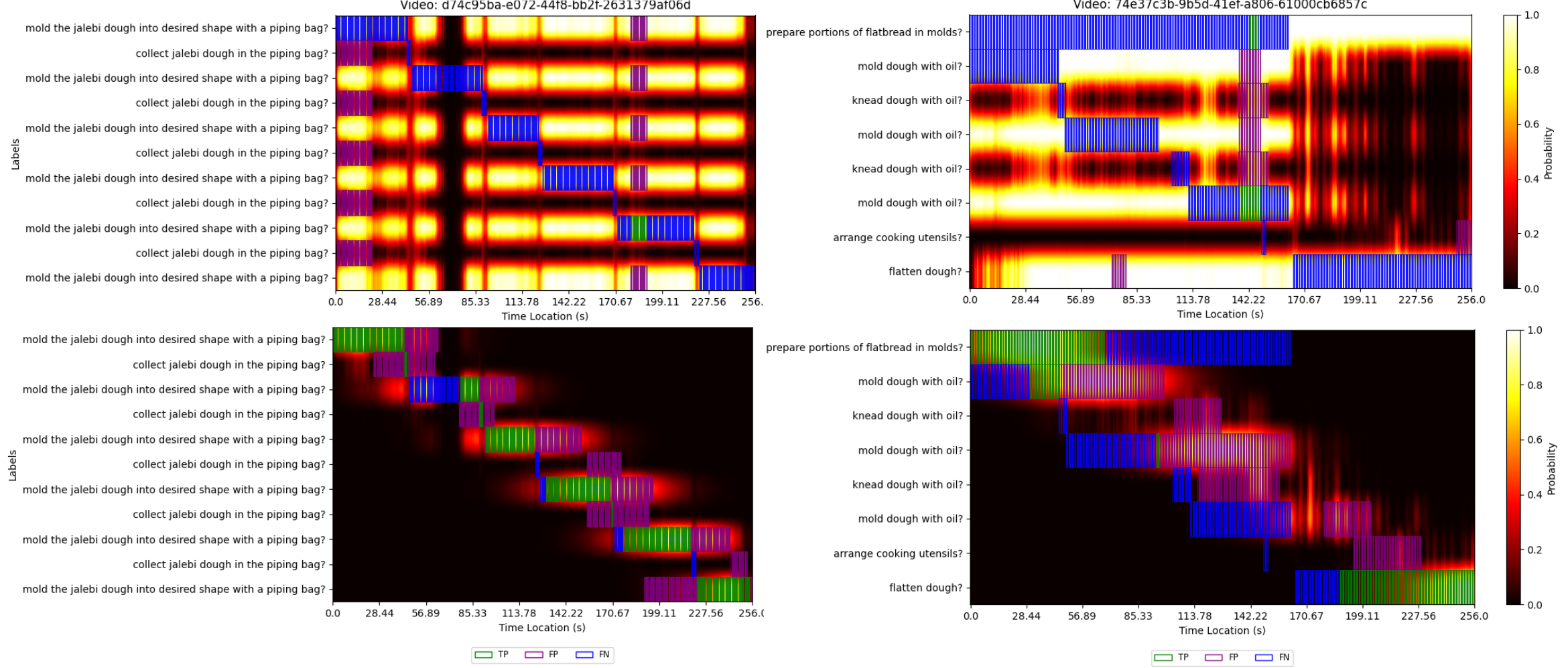} 


 \caption{\textbf{Positive (left) and negative (right) examples of our method before and after the temporal-order prior during the test time refinement.} We report the True Positives (TP) segments in green, the False Positives (FP) in purple, the False Negatives (FN) in blue and the probability score along the time location in number of segments.}
  \label{FigPrior1}
\end{figure*}

\paragraph{Qualitative results}

We show positive and negative examples of the qualtitative results on Figure \ref{FigPrior1}.
The results show the importance of the temporal-order to refine the probability scores, which excludes FP segments and centers the prediction according to its temporal order. However, since the VSLNet architecture adopts a sparse-sampling technique, the model struggles in short atomic actions due to the loss of fine-grained information when sampling.


\section{Conclusions}

The Bayesian VSLNet introduced in this work shows how to exploit the temporal order of the steps with the Bayesian rule test time refinement. However, it still suffers from the significant duration of the videos where a sparse-sampling technique is not enough. We hope that our paper inspires future work to identify and tackle the challenges of the Step Grounding task.
{
    \small
    \bibliographystyle{ieeenat_fullname}
    \bibliography{main}
}


\end{document}